\documentclass[letterpaper]{article} 
\usepackage{aaai24}  
\usepackage{times}  
\usepackage{helvet}  
\usepackage{courier}  
\usepackage[hyphens]{url}  
\usepackage{graphicx} 
\usepackage{threeparttable}
\usepackage{subcaption}
\usepackage{mwe}
\usepackage{bbding}
\usepackage{amsmath}
\usepackage{amssymb}
\usepackage{booktabs}
\usepackage{multirow}
\usepackage{colortbl}
\usepackage{xcolor}

\definecolor{lightergray}{gray}{0.9}

\usepackage[most]{tcolorbox}
\usepackage{lipsum}
\definecolor{linequote}{RGB}{224,215,188}
\definecolor{backquote}{RGB}{249,245,233}

\newtcolorbox{myquote}[1][]{%
    enhanced, breakable, 
    size=minimal,
    frame hidden, boxrule=0pt,
    sharp corners,
    colback=backquote,
    #1
}

\usepackage[T1]{fontenc}    %
\usepackage{mdframed}
\newmdenv[
  font=\ttfamily\small,
  font=\small,
  linewidth=0.5pt,
  innerleftmargin=10pt,
  innerrightmargin=10pt,
  innertopmargin=10pt,
  innerbottommargin=10pt,
]{monobox}

\usepackage{natbib}  
\usepackage{caption} 
\usepackage{algorithm}
\usepackage{algorithmic}

\usepackage{newfloat}
\usepackage{listings}
\DeclareCaptionStyle{ruled}{labelfont=normalfont,labelsep=colon,strut=off} 
\floatstyle{ruled}
\newfloat{listing}{tb}{lst}{}
\floatname{listing}{Listing}
\nocopyright 

\title{PMC-LLaMA: Towards Building Open-source Language Models for Medicine }
\author{
    Chaoyi Wu\textsuperscript{\rm 1,2,$\ast$},
    Weixiong Lin\textsuperscript{\rm 1,2,$\ast$},
    Xiaoman Zhang\textsuperscript{\rm 1,2},
    Ya Zhang\textsuperscript{\rm 1,2}\\
    Yanfeng Wang\textsuperscript{\rm 1,2},
    Weidi Xie\textsuperscript{\rm 1,2} 
}
\affiliations {
    \textsuperscript{\rm 1}Cooperative Medianet Innovation Center, Shanghai Jiao Tong University, Shanghai, China\\
    \textsuperscript{\rm 2}Shanghai AI Laboratory, Shanghai, China\\
    \{wtzxxxwcy02, wx\_lin, xm99sjtu, ya\_zhang, wangyanfeng, weidi\}@sjtu.edu.cn
}

\usepackage{bibentry}

\begin{document}

\maketitle

\begin{abstract}
Recently, Large Language Models (LLMs) have showcased remarkable capabilities in natural language understanding. While demonstrating proficiency in everyday conversations and question-answering situations, these models frequently struggle in domains that require precision, such as medical applications, due to their lack of domain-specific knowledge. In this paper, we describe the procedure for building a powerful, open-source language model specifically designed for medicine applications, termed as PMC-LLaMA. 
Our contributions are threefold: 
(i) we systematically investigate the process of adapting a general-purpose foundation language model towards medical domain, 
this involves data-centric knowledge injection through the integration of 4.8M biomedical academic papers and 30K medical textbooks, 
as well as comprehensive fine-tuning for alignment with domain-specific instructions; 
(ii) we contribute a large-scale, comprehensive dataset for instruction tuning. 
This dataset encompasses medical question-answering~(QA), rationale for reasoning, and conversational dialogues, comprising a total of 202M tokens;
(iii) we conduct thorough ablation studies to demonstrate the effectiveness of each proposed component. While evaluating on various public medical question-answering benchmarks, our lightweight PMC-LLaMA, which consists of only 13 billion parameters, exhibits superior performance, even surpassing ChatGPT. All models, codes, datasets can be found in \url{https://github.com/chaoyi-wu/PMC-LLaMA}.
\renewcommand{\thefootnote}{}
\footnotetext{$\ast$: Equal contributions.}

\end{abstract}

\section{Introduction}
\vspace{2pt}


The rapid advancement of large language models (LLMs), 
for example, OpenAI’s ChatGPT~\citep{chatgpt} and GPT-4~\citep{openai2023gpt4} has truly revolutionized the natural language processing research~\citep{nori2023capabilities,singhal2022large}, sparking AI applications for numerous daily scenarios.
Unfortunately, the training details and model architectures for the GPT-series remain unclear. 
The open-source LLMs, \emph{e.g.}, LLaMA-series~\citep{touvron2023llama, Touvron2023Llama2O}, also show comparable performance with ChatGPT in the general domain. However, though the LLMs demonstrate proficiency in everyday conversations, 
in medical domain where  requires high precision, they often produce seemingly accurate output but lead to incorrect conclusions, which could be highly fatal. 
We conjecture this is due to their lack of comprehensive medical knowledge. 

\begin{figure}
    \centering
    \includegraphics[width=\linewidth]{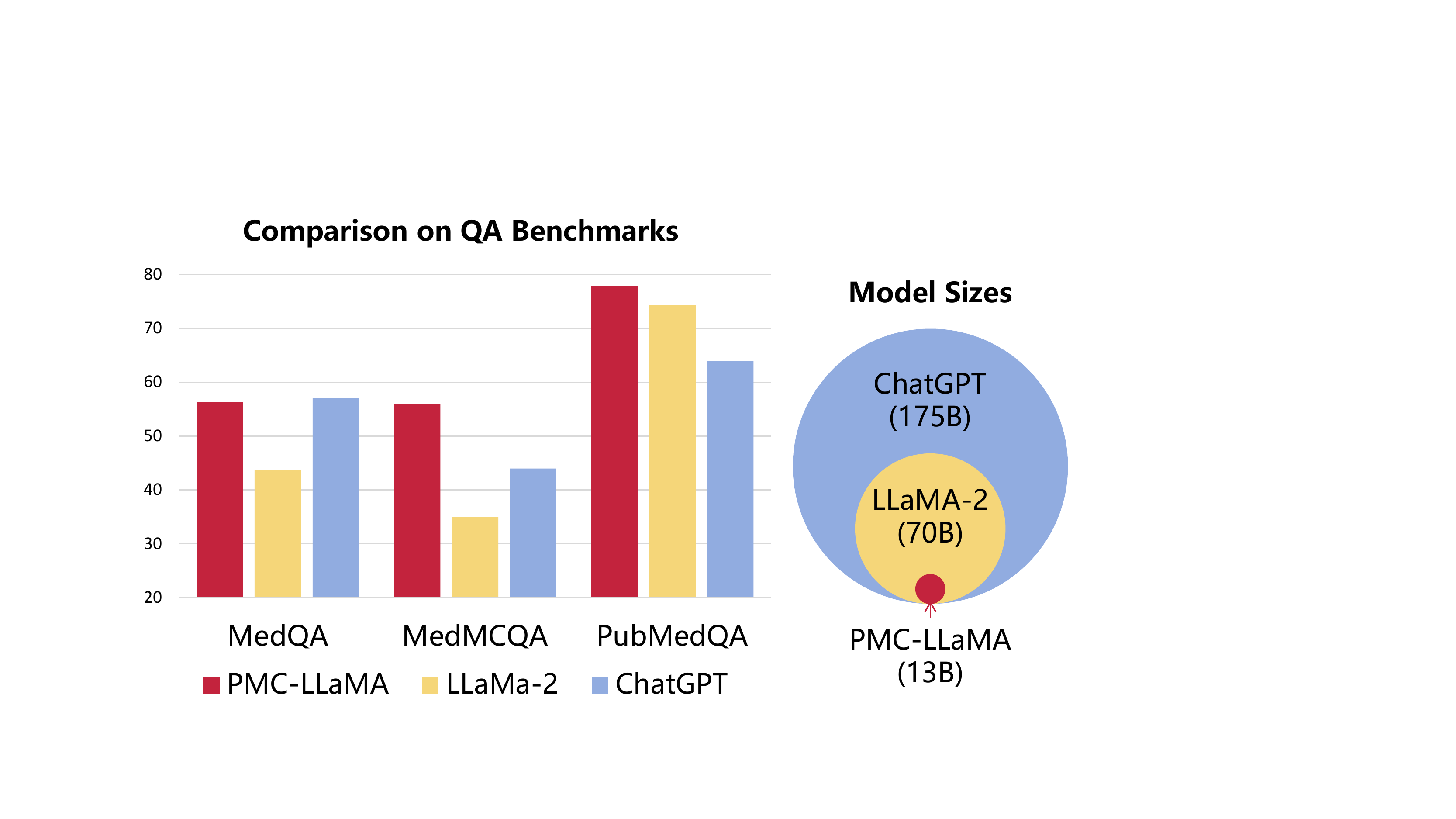}
    \vspace{-0.6cm}
    \caption{In the left, we show the general comparison between our PMC-LLaMA with LLaMA-2 and ChatGPT. On the right, we visually show the advantages of our model in model sizes. PMC-LLaMA is much smaller than the others.}
    \vspace{-0.4cm}
    \label{fig:begin}
\end{figure} 

Existing works have also explored several ways for adapting general-purpose LLMs towards
medicine domain, like Med-Alpaca~\cite{han2023medalpaca}, Chat-Doctor~\cite{yunxiang2023chatdoctor} and MedPALM-2~\cite{Anil2023PaLM2T}. 
Among these, MedPALM-2 is the only work successfully outperforming ChatGPT while their training details, for example, training data, model architecture, remain unclear. Thus, systematic investigation on the medical domain adaptation for LLMs still needs to be discussed further especially in open-source community.

Our goal is to systematically adapt an open-source general LLM, \emph{i.e.}, LLaMA, towards the medicine domain from the following aspects.  
First, we adopt data-centric  medical-specific knowledge injection for the language model with a large-scale free text medical corpora. We claim that language models can accumulate enough medical knowledge in this step and build up a better embedding space for domain-specific complex terminologies. 
Second, augmenting the reasoning capabilities of the proposed model.
This empowers the model to link its medical knowledge with provied case information and provide well-justified recommendations. 
Lastly, enhancing the alignment ability of LLMs.
Robust alignment with various instructions facilitates effective zero-shot adaptation to a diverse spectrum of tasks.




In conclusion, in this paper we systematically build up an LLM for medicine through data-centric knowledge injection and medical-specific instruction tuning, and release an open-source lightweight medical-specific language model, PMC-LLaMA. 
Specifically, we first collect a large medical-specific corpus, named MedC-K, consisting of \textbf{4.8M} biomedical academic papers and \textbf{30K} textbooks for knowledge injection. 
We then adopt medical-specific instruction tuning on a new medical knowledge-aware instruction dataset, termed MedC-I, consisting of medical QA, rationale, and conversation with \textbf{202M} tokens in total.
We evaluate PMC-LLaMA on various medical QA benchmarks, surpassing ChatGPT and LLaMA-2 as shown in Fig.~\ref{fig:begin}.

\section{Related Work}
\vspace{2pt}

\subsubsection{Large Language Model.}
Recently, the great success of large language models~(LLM)~\cite{chatgpt,openai2023gpt4, Anil2023PaLM2T,du2021glm}, 
has garnered significant attention within the field of natural language processing.
For example, OpenAI's strides with ChatGPT and GPT-4 have showcased remarkable capabilities in various tasks, including text generation, language translation, question answering, and more. However, intricate details concerning their training methodologies and weight parameters remain undisclosed. 
LLaMA~\cite{touvron2023llama} serves as an open-source alternative for the foundational language model, ranging from 7 billion to 65 billion parameters. 
In light of these advancements, there has been a surge of interest in tailoring language models for specific biomedical domains. Most of these models are prompt-tuned using LLaMA on a small medical corpus, resulting in a deficiency of comprehensive medical knowledge integration.

\subsubsection{Instruction Tuning.}
For LLMs to follow natural language instructions and complete real-world tasks, instruction-tuning has been widely used for alignment~\cite{ouyang2022training, peng2023instruction}. This involves fine-tuning the model on a collection of tasks described via instructions, to effectively improve the zero-shot and few-shot generalization abilities of LLMs~\cite{chung2022scaling, iyer2022opt}.
Building on the publicly accessible language models, 
Alpace~\cite{alpaca} and Vicuna~\cite{vicuna2023} are proposed, 
by finetuning on the machine-generated instruction-following samples, showing promising performance. In the medical domain, 
Chat-Doctor~\cite{yunxiang2023chatdoctor}, 
and Med-Alpaca~\cite{han2023medalpaca}, 
are instruction-tuned for medical question-answering and dialogue applications. Notably, Med-PaLM~\cite{singhal2022large} represents the pinnacle of LLMs in the medical field, trained with intensive instruction tuning on the strong PaLM model~(with 540 billion parameters). 
However, its code and data remain inaccessible to the public.


\subsubsection{Medical Foundational Language Model.}
In addition to instruction tuning, there has been extensive efforts on training foundation model for medicine, for example,
BioBert, BioMedGPT, {\em etc}.
~\cite{lee2020biobert,zhang2023biomedgpt,luo2022biogpt}.
However, these models exhibit certain limitations,
first, most domain-specific models have been exclusively trained on medical corpora.
The lack of exposure to diverse knowledge domains beyond medicine can impede the model's capability to perform reasoning or context understanding;
second, these models are limited in model scale and are predominantly designed to base on BERT, thus imposing restrictions on their utility for a wide array of downstream tasks under zero-shot learning.
In this work, we aim to resolve these two limitations by adapting a general LLM toward medicine with knowledge injection, followed by medical-specific instruction tuning.

\begin{figure*}[t]
    \centering
    \includegraphics[width=1\textwidth]{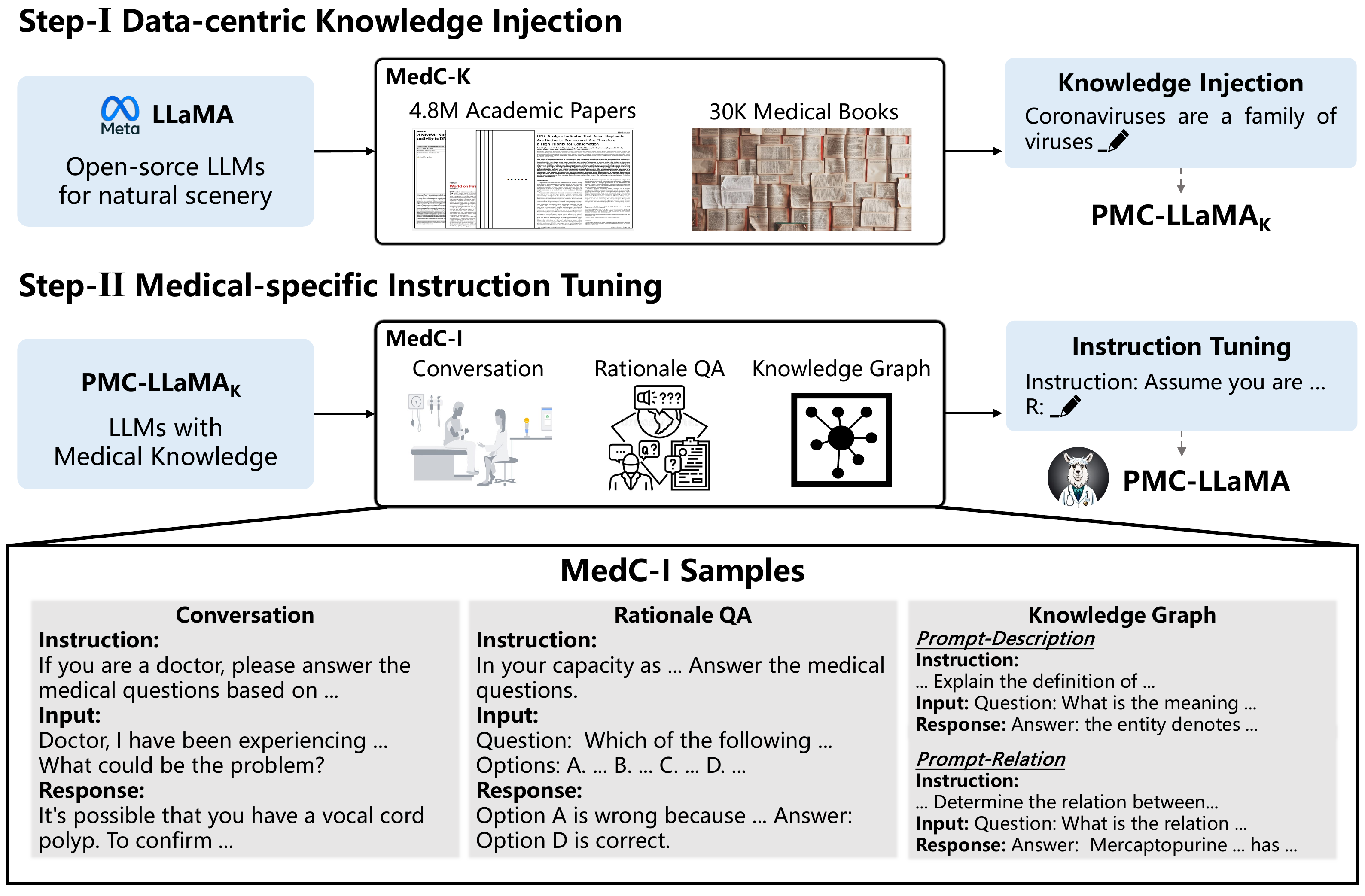}
    \caption{The training pipeline of PMC-LLaMA. Our training flow can be separated into two parts, \emph{i.e.}, data-centric knowledge injection and medical-specific instruction tuning. In knowledge injection, we collect 4.8M biomedical academic papers and 30K medical books for further injecting knowledge into LLaMA. In the instruction tuning stage, we mainly consider three aspects, medical conversation, medical rationale question-answering, and knowledge graph, containing 202M tokens in total.}
    \vspace{-8pt}
    \label{fig:teaser}
\end{figure*}

\section{Problem Formulation}
\vspace{2pt}

In this paper, our goal is to systematically investigate the procedure for steering a pre-trained foundational language model to the knowledge-intense domain, {\em i.e.}, medicine.
The training process can be divided into two stages: 
first, a data-centric knowledge injection stage, that aims to enrich the language model with fundamental medical knowledge; second, a medical-specific instruction tuning stage, that tailors the model to align with clinical use cases. 


At training stage, assuming the text input as a sequence of tokens, 
{\em e.g.}, $\mathcal{U} = \{u_1, u_2, \dots, u_N \}$, 
where each $u_i$ is a text token and $N$ is the total sequence length,
the training objective is to minimize auto-regressive loss, 
with the major difference on whether to compute loss on the entire sequence or only sub-sequence, as detailed in the following.

\subsubsection{Data-centric Knowledge Injection.} 
For the knowledge injection step, we simply minimize the default auto-regressive loss, all free-form texts on medical knowledge can be used, for the model to accumulate sufficient medical-specific knowledge contexts, formulated as
\begin{equation}
    L(\Phi) = -\sum \log \Phi(u_i|u_{\textless i}).
\end{equation}
where $ u_{\textless i}$ indicates the tokens appear before index $i$ and $\Phi$ denotes our model.

\subsubsection{Medical-specific Instruction Tuning.} 
At this stage, the token sequence is further split into instruction $\mathcal{I}$, and response $\mathcal{R}$, the former is to mimic user's query, thus the loss is ignored at training time, denoted as:
\begin{equation}
    L(\Phi) = -\sum_{u_i \in \mathcal{R}} \log \Phi(u_i|u_{\textless i},\mathcal{I}).
\end{equation}
At inference time, the common use case is a conversation, 
where the user normally provides the question as instruction $\mathcal{I}$, 
and the output of the model serves as the answer.

\section{Dataset Construction}
To support our two-stage training, namely data-centric knowledge injection, and medical-specific instruction tuning for alignment, we herein detail the procedure for constructing the high-quality language datasets.

\subsection{Dataset-I: Fundamental Medical Knowledge} \vspace{1pt}
To steer a general-purpose foundational language model for medical scenario, 
we propose to first conduct data-centric knowledge injection, 
that aims to expose the model with medical-related terminologies and definitions.
We primarily focus on two key data sources, namely, biomedical papers and textbooks.

\subsubsection{Papers.} 
As a valuable knowledge resource, 
academic papers naturally contains high-quality, cutting-edge medical knowledge. 
We start with the S2ORC~\cite{lo-wang-2020-s2orc} Datasets with 81.1M English-language academic papers, and pick out those biomedical related papers depending on whether having corresponding PubMed Central~(PMC) IDs. As a result, there are around 4.8M biomedical papers left, totaling over 75B tokens.

\subsubsection{Books.} 
\begin{figure*}[t]
    \centering
    \includegraphics[width=1\textwidth]{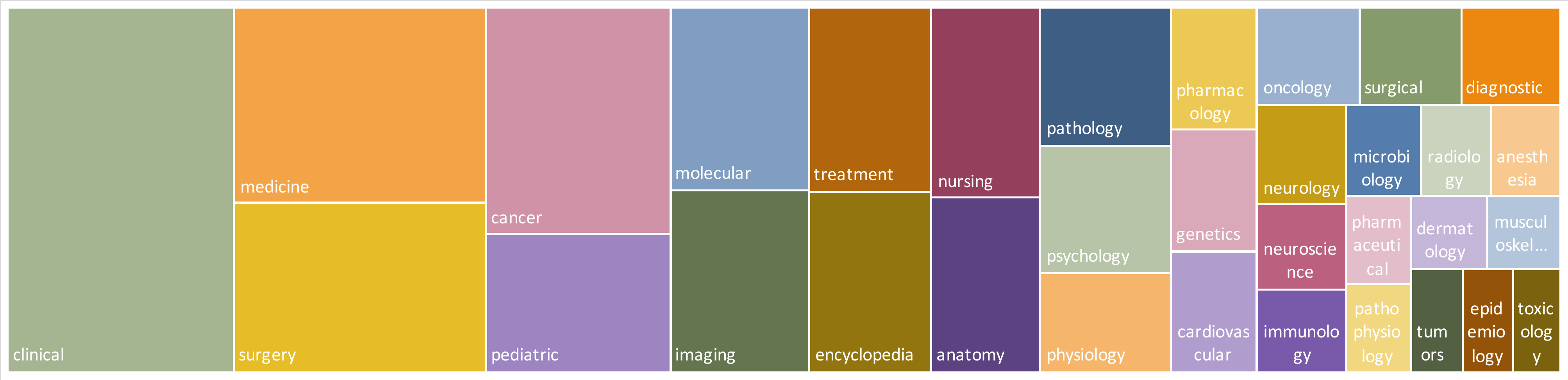}
    \caption{Distribution of medical textbooks categories. 
    The box sizes denote the book numbers for different categories.}
    \vspace{-13pt}
    \label{fig:data}
\end{figure*}

We collect 30K textbooks sourced from various outlets, 
for example, the open-library, university library, and reputable publishers, covering a wide range of medical specialties as shown in Fig.~\ref{fig:data}. 
For preprocessing, we first extract the text content from the book PDF, 
then carry out data cleaning via de-duplication and content filtering. 
Specifically, we eliminate extraneous elements such as URLs, author lists, superfluous information, document contents, references, and citations. 
Additionally, we have also removed any references to images and tables within the paragraphs, for example, `Fig. 1'. After this thorough cleaning process, 
there are approximately 4B tokens left.

\subsubsection{Combination.} 
The two corpora encompass distinct types of medical knowledge, 
while papers predominantly capture cutting-edge insights, 
books capture more fundamental medical knowledge, 
which is more crucial for pre-trained general-purpose language models.
Hence, when blending these two datasets for knowledge injection training,
we use a ratio of 15:4:1 at each training batch, by that we mean to emphasize ``book'' tokens more. Specifically, we sample more tokens from books, ensuring they occupy 15 parts per batch and sample tokens from ``papers'' less so that they occupy 4 parts per batch. For the remaining 1 occupation, we sample from a general language corpus, RedPajama-Data~\cite{together2023redpajama} to form a complete batch. This mainly aims to avert catastrophic forgetting of previously acquired general text knowledge after extensive knowledge injection on large-scale medical-specific data.

\subsubsection{Knowledge Injection Training.} 
Till here, we have constructed a large-scale language dataset of fundamental medical knowledge, termed as \textbf{MedC-K}. With such corpus, 
we conduct data-centric knowledge injection with auto-regressive training, 
resulting a language model for medicine, named as \textbf{PMC-LLaMA$_\text{K}$}, 
as the largest number of tokens are from PubMed Central academic papers.



\subsection{Dataset-II: Medical Instructions}
Here, we proceed to carry out instruction tuning with the goal of guiding the model to respond to various instructions, by exploiting the medical knowledge embedded in PMC-LLaMA$_\text{K}$ model. Generally speaking, our instruction tuning datasets are composed of three main parts, namely, medical consulting conversation, 
medical rationale QA, and medical knowledge graph prompting. 

\subsubsection{Medical Conversation.}
Considering there exists diverse doctor-patient dialogues in daily life, 
the questions raised by patients are naturally suitable as instructions and doctor responses as ground truth. We start with the data collected by Med-Alpaca~\cite{han2023medalpaca} and Chat-Doctor~\cite{yunxiang2023chatdoctor}, and further expand the provided instructions into various synonymous sentences to improve model's robustness to diverse instructions. Specifically, we use the GPT-4 with the following query prompt:

\begin{myquote}[colback=gray!30, size=small]
    ``Rewrite 10 sentences that convey similar meanings to what I've stated: \{instruction seeds\}.'',
\end{myquote}

\noindent where \{instruction seeds\} denotes the provided instruction from ChatDoctor or MedAlpaca, and the query can be repeated until the desired prompt number. At training time, we randomly select one instruction from the instruction base, to simulate the inputs from real users and avoid over-fitting on specific instruction templates.

\subsubsection{Medical Rationale QA.} 
Beyond daily conversations, we also consider equipping our model with reasoning ability with professional medical knowledge. 
We start with the training sets of the open-source medical multi-choice question-answering datasets,
such as USMLE~\cite{jin2021disease}, PubMedQA~\cite{jin2019pubmedqa} and MedMCQA~\cite{pal2022medmcqa}. 
Despite the questions in them naturally demanding medical-specific knowledge, 
most of these datasets only include plain choices, lacking detailed reasoning guidance. To complement such information, we prompt ChatGPT~\cite{chatgpt} for causality analysis. 
Specifically, given a QA pair, we query ChatGPT to get rationale output~(check supplementary for details), and treat the output as an explanation with structured format shown at the bottom of Fig.~\ref{fig:teaser}.



    


\subsubsection{Medical Knowledge Graph Prompting.}
In addition to the aforementioned data, 
we also consider exploiting medical knowledge graphs UMLS~\cite{lindberg1993unified}, to align with clinicians' experience.
Specifically, to link the medical terminologies with their respective knowledge description or corresponding relationships, we construct QA pairs to translate the common knowledge graph. There are two main types contained in medical knowledge graph, \emph{i.e.}, entity descriptions and entity relationships. We add two different prompts for them as shown at the bottom of Fig.~\ref{fig:teaser}, that demands the model to output descriptions for a certain entity or predict the relationship between two entities.

\subsubsection{Medical-spcific Instruction Tuning.} 
By combining the above three parts together, 
we form a large-scale, high-quality, medical-specific instruction tuning dataset, \textbf{MedC-I}, consisting \textbf{202M} tokens. 
We further tune PMC-LLaMA$_\text{K}$ on it, 
resulting in our final model -- \textbf{PMC-LLaMA}.

\section{Experiment}
\subsection{Training Details}
We start by carrying out knowledge injection on open-source LLaMA model, 
optimizing an auto-regressive loss. Specifically, at training time, 
the max context length is set as 2048, with a batch size to be 3200, 
and the model is trained with AdamW optimizer~\citep{loshchilov2017decoupled} 
with a learning rate 2e-5. We adopt the Fully Sharded Data Parallel~(FSDP) acceleration strategy, bf16~(Brain Floating Point) data format, and gradient checkpointing~\cite{chen2016training}. Since we sample more tokens from books in each batch, the model will finish seeing all book tokens earlier. Thus, we here define 1 epoch for seeing all book tokens instead of seeing all mixed tokens. The model is trained with knowledge injection for 5 epochs with 32 A100 GPUs. Then we carry out medical-specific instruction tuning on MedC-I, for 3 epochs with 256 batch sizewith 8 A100 GPUs. Note that, at instruction tuning stage, each epoch refers to looping through all sequences.

\subsection{Benchmarks}
In the literature, the primary method for measuring the ability of medical language models is based on multiple-choice question answering, which uses accuracy as the main metric. Following the convention, we adopt three prominent medical question-answering (QA) benchmarks for evaluation. 

\begin{itemize}
\setlength\itemsep{2pt}
    \item PubMedQA~\cite{jin2019pubmedqa} is a biomedical QA dataset collected from PubMed abstracts.  The task of PubMedQA is to answer research questions with yes/no/maybe, which can be considered as the multiple-choice question.
    It is split into three subsets: 1k manually labeled pairs~(PQA-L), 61.2k unlabeled pairs~(PQA-U), and 211.3k artificially generated pairs~(PQA-A).
    Following former works~\cite{lmflow}, we view PQA-A as the train set, PQA-L as the test set, and discard the PQA-U parts.
    
    \item MedMCQA~\cite{pal2022medmcqa} is a dataset of multiple choice questions, that are sourced from mock exams and past exams of two Indian medical school entrance exams called AIIMS and NEET-PG~\citep{pal2022medmcqa}. 
    The train split contains 182,822 questions, and the test split contains 4183 questions. Each question has 4 choices.

    \item USMLE~\cite{jin2021disease} is a dataset of multiple choice questions~(4 choices per question), based on the United States Medical License Exams. 
    The dataset is collected from the professional medical board exams, 
    covering three languages: English, simplified Chinese, and traditional Chinese, containing 12,724, 34,251, and 14,123 questions respectively. 
    Here, we use the English parts and split it into 10,178 questions for training, 1273 for validation, and 1273 for testing, following the official splits.
\end{itemize}

\subsection{Baseline Models}
\subsubsection{LLaMA~\cite{touvron2023llama}.} 
LLaMA is the most widely-used open-source language model, it has been trained on a large text corpus with only auto-regressive learning, 
{\em i.e.}, no instruction tuning is involved.

\vspace{-1pt}
\subsubsection{LLaMA-2~\cite{Touvron2023Llama2O}.}
LLaMA-2 is the improved version of LLaMA that has been further tuned with instructions. Its largest version~(70B) is reported to be the best on natural scenery among the open-source LLMs.

\vspace{-1pt}
\subsubsection{ChatGPT~\cite{chatgpt}.} 
ChatGPT is a commercial model released by OpenAI in November, 2022, that has shown remarkable performance on a wide range of NLP tasks in various domains, including medicine. Note that, since the exact details of ChatGPT are confidential, 
we follow the general presumption that ChatGPT is roughly the same as GPT-3 in model sizes~(175B)~\cite{Kung2022PerformanceOC}.

\vspace{-1pt}
\subsubsection{Med-Alpaca~\cite{han2023medalpaca}.}
Med-Alpaca is a model further fine-tuned on Alpaca~\cite{alpaca} using medical instruction data. They focus on the task of assisting medical dialogues and question-answering.

\vspace{-1pt}
\subsubsection{Chat-Doctor~\cite{yunxiang2023chatdoctor}.}
Chat-Doctor is a language model aiming for health assistants, 
that is designed to provide users with medical information, advice, and guidance. 
For training, it has leveraged the dialogue-based instruction tuning data.

\subsection{Evaluation Settings}
\label{sec:eval}

In this section, we describe the evaluating detail to compare the above language models on the QA benchmarks. {\bf Note that}, we do not claim the presented comparison to be completely fair, as a number of training details, 
for example, data, architecture remain undisclosed for the commercial model. 
Therefore, we only treat these baseline models for reference, and more focused on presenting our procedure for building on a powerful language model for medicine.

Our evaluation settings can be divided into two types: task-specific fine-tuning evaluation and zero-shot instruction evaluation. 

\vspace{-1pt}
\subsubsection{Task-specific Fine-tuning Evaluation.} 
In this evaluation setting, we use the combination of three QA training sets to further fine-tune a language model and then evaluate it. For models without instruction tuning, for example, LLaMA and PMC-LLaMA$_\text{K}$, 
we adopt this evaluation setting by default.

\vspace{-1pt}
\subsubsection{Zero-shot Instruction Evaluation.} 
In this evaluation setting, we directly test the model by giving a medical QA instruction, {\em e.g.}, ``Make a choice based on the question and options.'', 
without doing any task-specific fine-tuning. 
Most models are evaluated in this setting, 
\emph{i.e.}, LLaMA-2, Med-Alpaca, Chat-Doctor, ChatGPT, and our own PMC-LLaMA.

\begin{table*}[htb!]
    \centering
    \footnotesize
    \setlength{\tabcolsep}{2pt} 
    \renewcommand{\arraystretch}{0.95} 
    \caption{Ablation study on QA benchmarks. ACC scores are reported in the table. Note that for the models without ability to follow instruction, we task-specific fine-tune them on the combination of the three downstream training sets to get the number.}
    \begin{tabular}{c|c|cc|ccc|ccc}
    \toprule
    \multirow{2}{*}{Method} & \multirow{2}{*}{Model Size} & \multicolumn{2}{c|}{Knowledge Injection} & \multicolumn{3}{c|}{Instruction Tuning}     & \multirow{2}{*}{MedQA} & \multirow{2}{*}{MedMCQA} & \multirow{2}{*}{PubMedQA} \\
                            &                             & Papers         & Books        & Rationale & Conversation & Knowledge Graph &       &         &          \\ \midrule
    Baseline~(LLaMA)  & 7B & \XSolidBrush & \XSolidBrush & \XSolidBrush & \XSolidBrush &  \XSolidBrush & 44.54 & 48.51  & 73.40\\ 
    Baseline~(LLaMA)  & 13B & \XSolidBrush & \XSolidBrush & \XSolidBrush & \XSolidBrush &  \XSolidBrush & 45.48 & 51.42  & 76.40\\
    \midrule
    \multirow{3}{*}{PMC-LLaMA$_\text{K}$} & 7B  & \Checkmark & \XSolidBrush & \XSolidBrush&\XSolidBrush &\XSolidBrush & 44.70 & 50.54 & 69.50 \\ 
         & 7B & \Checkmark & \Checkmark &\XSolidBrush &\XSolidBrush & \XSolidBrush& 45.56 & 51.45 & 74.60 \\
         & 13B & \Checkmark & \Checkmark  &\XSolidBrush &\XSolidBrush & \XSolidBrush& 48.15 & 54.15  & 77.10\\ \midrule
    \multirow{3}{*}{PMC-LLaMA} & 13B & \Checkmark & \Checkmark & \Checkmark &\XSolidBrush & \XSolidBrush & 49.32 & 54.56 & 77.20 \\
         & 13B & \Checkmark & \Checkmark & \Checkmark &  \Checkmark& \XSolidBrush& 54.43 & 55.77  & 77.00 \\
         & 13B&  \Checkmark &  \Checkmark & \Checkmark&  \Checkmark&  \Checkmark& \textbf{56.36} & \textbf{56.04} & \textbf{77.90} \\ \bottomrule
    \end{tabular}
    \label{tab:ablation}
\end{table*}

\begin{table*}[htb]
    \centering
    \footnotesize
    \setlength{\tabcolsep}{14pt} 
    \renewcommand{\arraystretch}{1.2} 
    \caption{Evaluation on QA Benchmarks. ACC scores are reported. Average refers to the average of the three datasets.}
    \vspace{-5pt}
    \begin{tabular}{l|c|ccc| >{\columncolor{lightergray}}c}
    \toprule
        Methods & Model Size & MedQA & MedMCQA & PubMedQA & Average \\
\toprule
        Human~(pass) & - & 50.0 & - & 60.0  &  - \\
        Human~(expert) & - & 87.0 & 90.0 & 78.0 & 85.0 \\
        \midrule
        ChatGPT~\cite{chatgpt} & 175B & \textbf{57.0} & 44.0 & 63.9 & 54.97 \\
        LLaMA-2~\cite{Touvron2023Llama2O}  & 13B & 42.73 & 37.41  & 68.0 & 49.40 \\
        LLaMA-2~\cite{Touvron2023Llama2O}   & 70B & 43.68 & 35.02 & 74.3 & 51.00 \\
        Med-Alpaca~\cite{han2023medalpaca} & 13B & 30.85 & 31.13 & 53.2 & 38.38 \\
        Chat-Doctor~\cite{yunxiang2023chatdoctor} & 7B & 33.93 & 31.10 & 54.3 & 39.78\\
        \midrule
        PMC-LLaMA & 13B & \textbf{56.36} & \textbf{56.04} & \textbf{77.9} & \textbf{64.43} \\
        \bottomrule
    \end{tabular}
    \label{tab:eval_on_qa_benchmark}
\end{table*}

\begin{figure*}[htb]
\centering
\begin{subfigure}[][][t]{0.33\textwidth}
    \includegraphics[width=\textwidth]{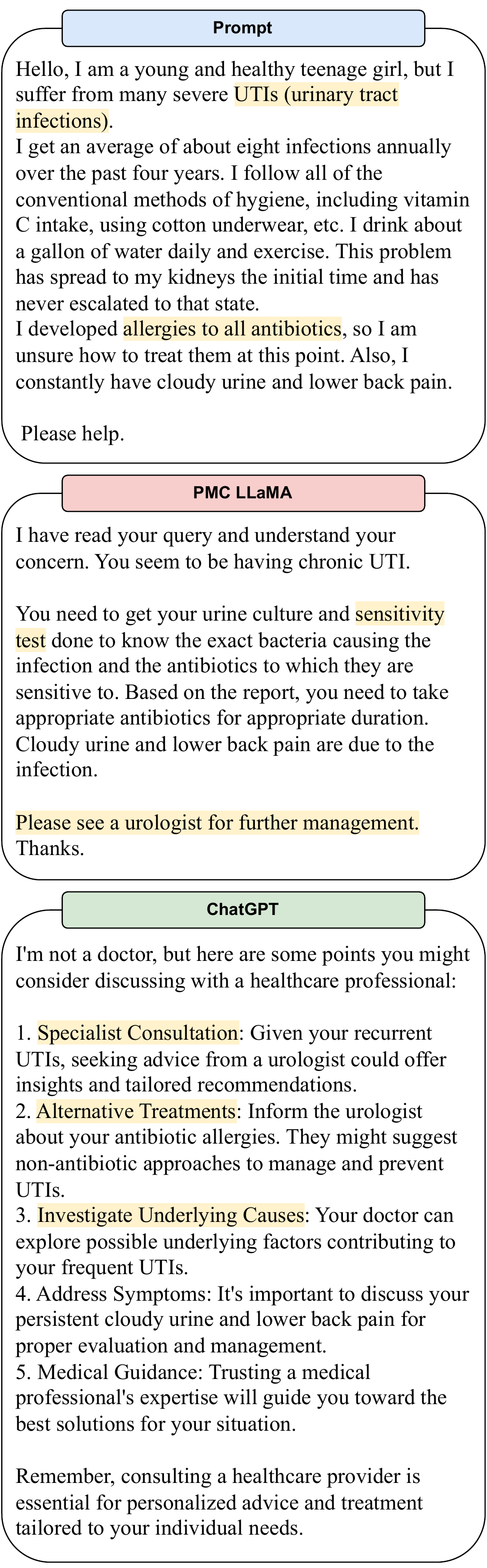}
    \caption{Patient-Physician Conversation}
    \label{fig:case_chat}
\end{subfigure}
\hfill
\begin{subfigure}[][][t]{0.33\textwidth}
    \includegraphics[width=\textwidth]{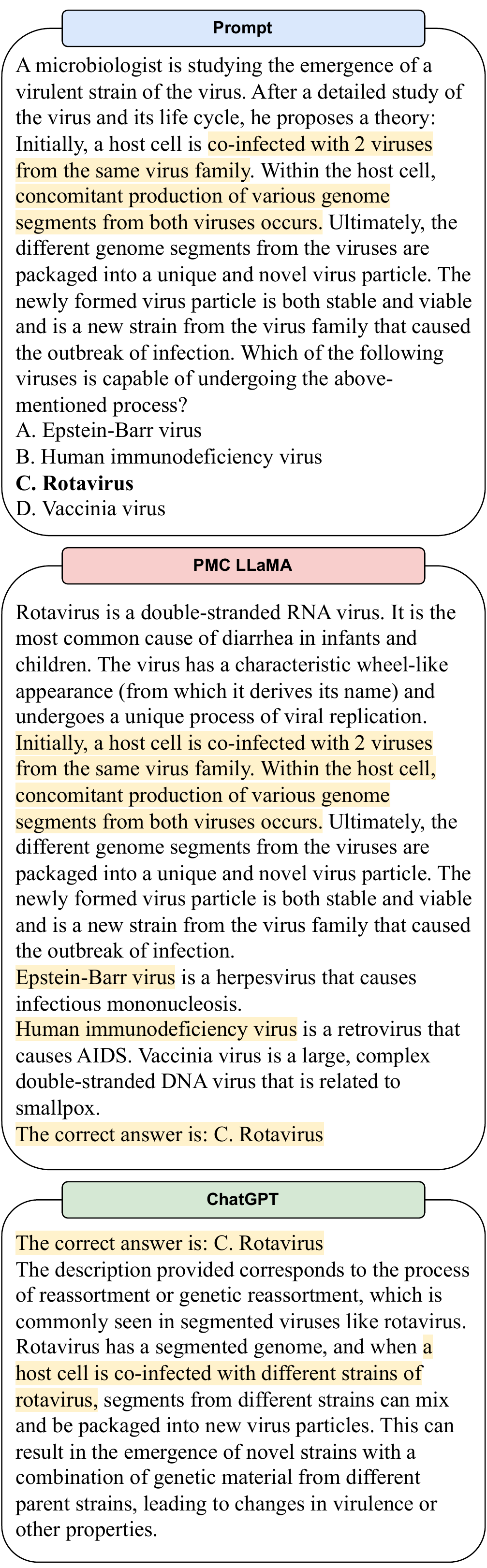}
    \caption{Microbiology Question}
    \label{fig:case_rationale_1}
\end{subfigure}
\hfill
\begin{subfigure}[][][t]{0.33\textwidth}
    \includegraphics[width=\textwidth]{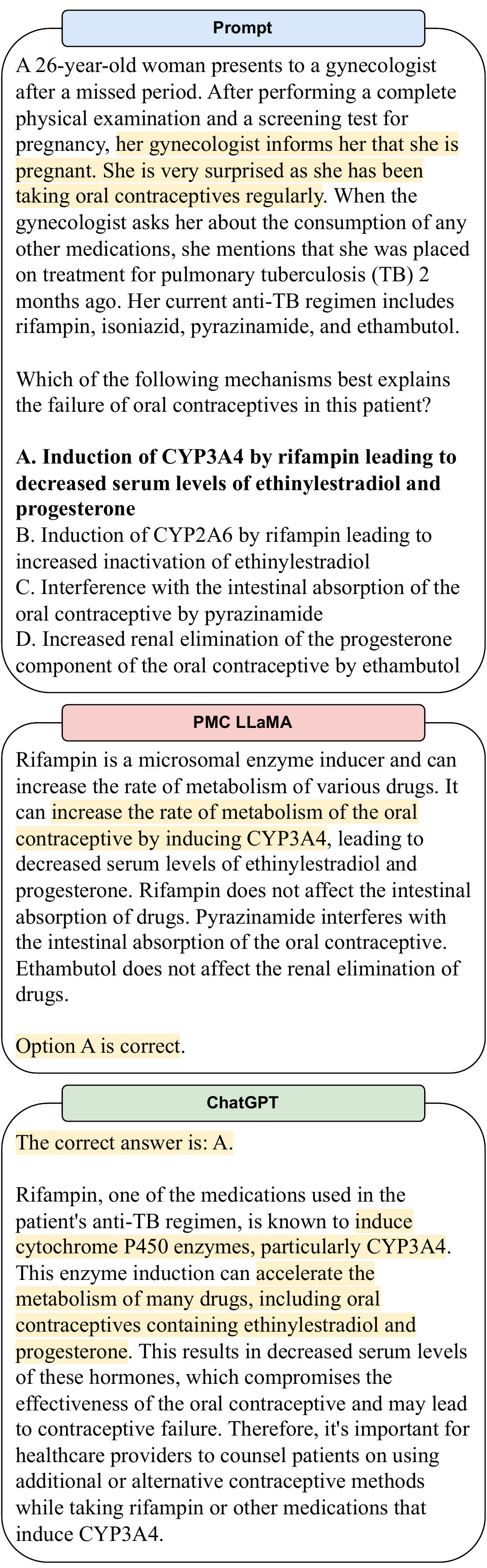}
    \caption{Pharmacology Question}
    \label{fig:case_rationale_2}
\end{subfigure}
\caption{Examples of three zero-shot cases from PMC-LLaMA and ChatGPT. (a) compares their responses to the patient's query, where PMC-LLaMA proposes more concrete suggestions. (b) shows the probing of microbiology knowledge. And PMC-LLaMA analyze both correct and incorrect options, enhancing the comprehensiveness of the analysis.
Example (c) examines the models' grasp of Pharmacology, and they respond with roughly equivalent answers.
The correct options are marked \textbf{bold}.
}
\label{fig:cases}
\end{figure*}

\section{Results}

\vspace{2pt}

In this section, we will introduce the experimental results. 
First, we conduct thorough ablation study on medical QA benchmarks, 
to demonstrate the effectiveness of the different components in our training procedure.
Then we show the comparison with different SOTA methods. Lastly, we present qualitative cases studies.




\subsection{Ablation Study}
As shown in Tab.~\ref{tab:ablation}, we systematically study the different design choices on various medical QA benchmarks, for example, effect of the model scale, data-centric knowledge injection, and medical-specific instruction tuning.

\vspace{-2pt}
\subsubsection{Model scale.} 
The scaling law~\cite{Kaplan2020ScalingLF} can also be observed in the medical corpus, for example, as shown in the table, when switching the model size from 7B to 13B, performance on all benchmarks have been improved. This phenomenon holds for both baseline LLaMA model and PMC-LLaMA$_\text{K}$, which has further trained with fundamental medical knowledge.


\vspace{-2pt}
\subsubsection{Data-centric knowledge injection.}
Compared with baseline 7B LLaMA model, integrating biomedical papers brings a performance gain from $44.54\%$ to $44.70\%$ and $48.51\%$ to $50.54\%$ on MedQA and MedMCQA respectively. While after adding books for training, the performance is improved significantly, \emph{i.e.}, obtaining $1.02\%$, $2.94\%$, and $1.2\%$ on MedQA, MedMCQA and PubMedQA respectively. Both observations have shown the importance of injecting fundamental medical knowledge.

\subsubsection{Medical-specific instruction tuning.}
We start instruction tuning with only rationale QA data. 
In this cases, since only QA task is considered, 
the difference from task-specific fine-tuning only lies on whether to give rationale sentence as supervision signal. We observe that simply incorporating rationale cases can lead to enhance QA results compared to task-specific fine-tuning on plain choice data, showcasing an improvement of 1.17\% on the MedQA dataset. 



Furthermore, integrating conversations with rationale QA for instruction tuning can produce substantial enhancements, with performance boosts from $49.32\%$ to $54.43\%$ on MedQA. This demonstrates the pivotal role played by the diversity of question types during the instruction tuning stage, as all involved questions will be limited on medical choice tests without conversation.
In addition, the incorporation of a knowledge graph introduces a further improvement of 1.93\% on the MedQA dataset, demonstrating the importance of using explicit instructions to emphasize the key medical concepts.


\subsection{Comparison with Baselines}
In Tab.~\ref{tab:eval_on_qa_benchmark}, we conduct a comparative analysis of our model against SOTA baseline models on three QA benchmark datasets for evaluation.
We also show a qualitative case study to demonstrate the conversation and rationale ability.

\subsubsection{Medical QA Ability.}
While comparing with other large language models on medical QA benchmarks, PMC-LLaMA achieves superior results on most of them, improving the average accuracy from $54.97\%$ to $64.43\%$, even surpassing the powerful ChatGPT,
despite containing significantly fewer parameters.

\subsubsection{Zero-shot Case Study.}
In Fig.~\ref{fig:cases}, we show qualitative examples with the zero-shot prediction from PMC-LLaMA and ChatGPT to verify the quality of prediction, 
covering patient-physician conversation and rationale QA.
The query in Fig.~\ref{fig:case_chat} is raised online after our data collection, thus, none of the models have seen this the question at training time. 
Based on the patient's description, both PMC-LLaMA and ChatGPT recognize the symptom of recurrent UTIs~(urinary tract infections), while PMC-LLaMA proposes a sensitivity test as the specific advice, rather than the general suggestion~(investigate the underlying causes) given by ChatGPT.
Fig.~\ref{fig:case_rationale_1} shows a QA case of microbiology. 
As can be seen, PMC-LLaMA not only produces the accurate answer, 
but also briefly analyzes the wrong options, forming a more comprehensive rationale.
Another case that focuses on pharmacology knowledge is illustrated in Fig.~\ref{fig:case_rationale_2}. Both PMC-LLaMA and ChatGPT have shown to properly understand Rifampin's efficacy and mechanism of side effects.

\section{Conclusion}
\vspace{2pt}

In this paper, we have systematically investigated the procedure for building up a medical-specific large language model based on an open-source large language model, including data-centric knowledge injection and medical-specific instruction tuning.
As a result, our proposed PMC-LLaMA is the first, open-source medical-specific language model, that demonstrates superior performance on various medical benchmarks, surpassing ChatGPT and LLaMA-2 with much fewer parameters.


\clearpage
\bibliography{aaai24}

\newpage
\section{Supplementary}

In addition to previously introduced methods, we provide detailed implementation of prompt templates used to query ChatGPT to form our instruction tuning data. Besides, we also demonstrate more few-shot samples comparing PMC-LLaMA and ChatGPT.

\subsection{Prompt ChatGPT for Rationale QA}
To form a rationale QA dataset, we have constructed several query prompts to distillate the response of ChatGPT. In this section we show details about the main two types of query prompts we use. Based on their response format, they can be divided into \textbf{general-wise rationale prompts} and \textbf{optional-wise rationale prompts}. In both cases, 
ChatGPT will be presented with QA pairs and solicited to generate rationales for gap mitigation from the question to answer.

\paragraph{General-wise Rationale Prompts.} 
\begin{monobox}

\noindent\textcolor{gray}{\# Instruction}\\
Provide analysis about the question, take the following two questions as examples\\

\noindent\textcolor{gray}{\# Few-shot Example 1}

\noindent \textbf{Quesion}:
Chronic urethral obstruction due to benign prismatic hyperplasia can lead to the following change in kidney parenchyma

\noindent
A. Hyperplasia\\
B. Hyperophy\\
C. Atrophy\\
D. Dyplasia\\

\noindent
The answer is Option C Atrophy, so the analysis is
Chronic urethral obstruction because of urinary calculi, prostatic hyperophy, tumors, normal pregnancy, tumors, uterine prolapse or functional disorders cause hydronephrosis which by definition is used to describe dilatation of renal pelvis and calculus associated with progressive atrophy of the kidney due to obstruction to the outflow of urine.\\

\noindent\textcolor{gray}{\# Few-shot Example 2}

\noindent \textbf{Quesion}:
Which vitamin is supplied from only animal source?

\noindent
A. Vitamin C\\
B. Vitamin B7\\
C. Vitamin B12\\
D. Vitamin D\\

\noindent
The answer is Option C Vitamin B12, so the analysis is
Vitamin B12 (Cobalamin) is synthesized solely by microorganisms. In humans, the only source for humans is food of animal origin, e.g., meat, fish, and dairy products. Vegetables, fruits, and other foods of nonanimal origin doesn't contain Vitamin B12 . Daily requirements of vitamin Bp is about 1-3 pg. Body stores are of the order of 2-3 mg, sufficient for 3-4 years if supplies are completely cut off.\\

\noindent
\textbf{Now help me with another question}

\noindent
\{new question\}\\
The answer is \{answer idx\}, so the analysis is

\end{monobox}

\noindent In the example prompt upper, we provide two few-shot examples, guiding ChatGPT to generate comprehensive rationales for the question. Through this prompt query, the response will be organized in a whole sentence.

\paragraph{Optional-wise Rationale Prompts} 
\begin{monobox}

\{the quesion\}

\noindent
Answer: \{answer idx\}

\noindent
Analyze each option in detail in the format of\\
Option A is TRUE. [option analysis for A]\\
Option B is FALSE. [option analysis for B]\\
Option C is FALSE. [option analysis for C]\\
Option D is FALSE. [option analysis for D]

\end{monobox}

\noindent In the example prompt upper, we provide an examples to guide ChatGPT to generate option-wise rationale. We hope the answer can be analysed per option so that mode dense analysis can be added for model knowledge guidance.

\subsection{Zero-shot Samples}
Due to the space limitaion in the main body, we only show three cases. Here we present a broader array of zero-shot cases about conversation and rationale question-answering, providing a more encompassing perspective of PMC-LLaMA's capabilities.\\

\begin{figure*}[htb]
\centering
\begin{subfigure}[][][t]{0.33\textwidth}
    \includegraphics[width=\textwidth]{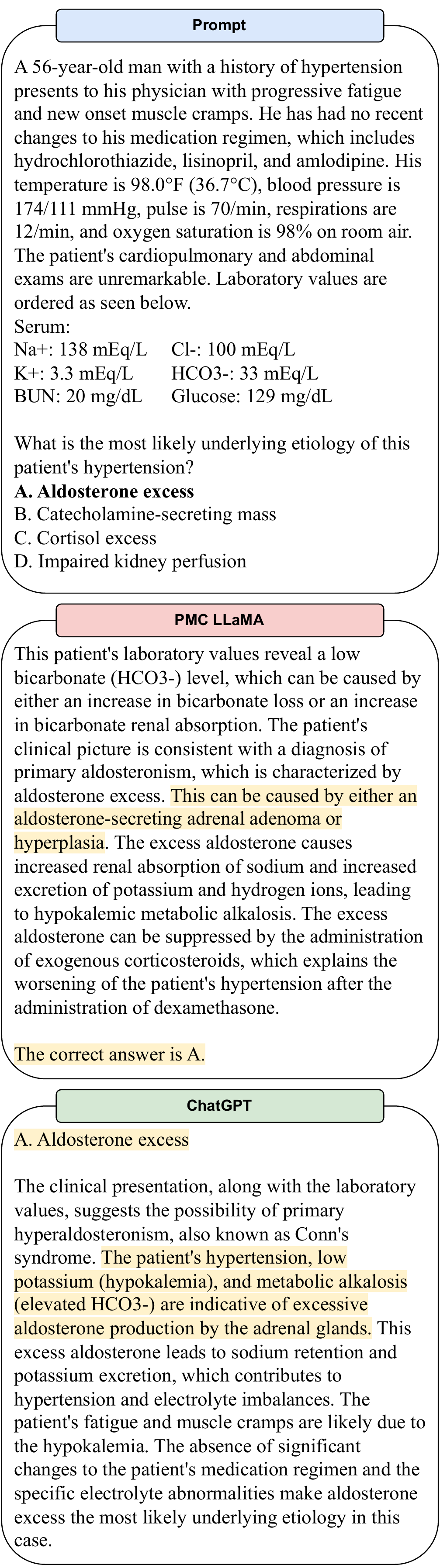}
    \caption{Neurology}
    \label{fig:case_chat}
\end{subfigure}
\hfill
\begin{subfigure}[][][t]{0.33\textwidth}
    \includegraphics[width=\textwidth]{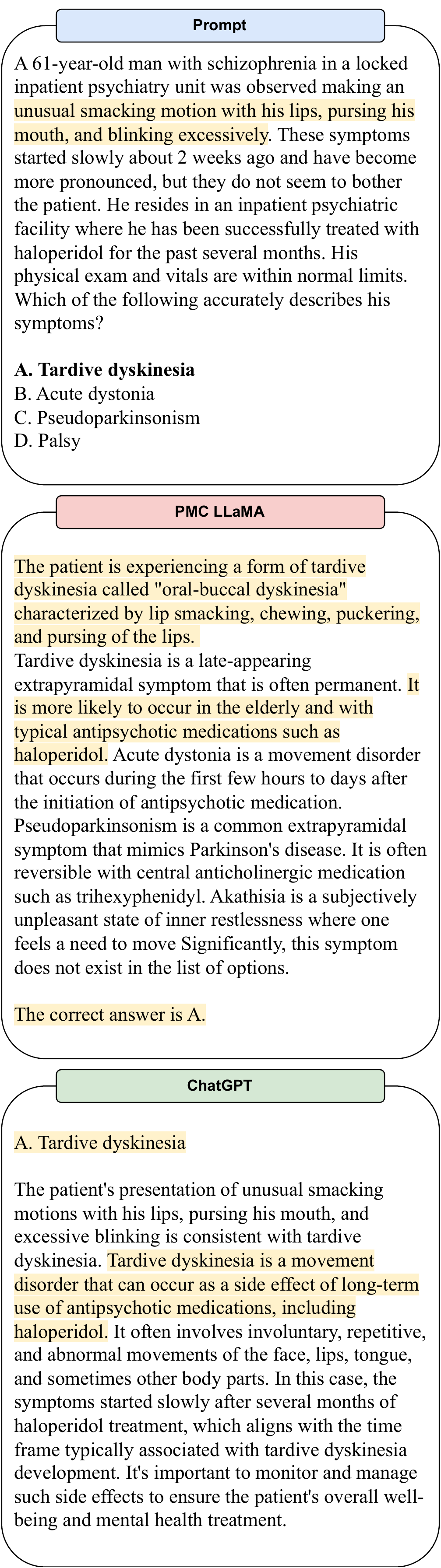}
    \caption{Psychiatry}
    \label{fig:case_rationale_1}
\end{subfigure}
\hfill
\begin{subfigure}[][][t]{0.33\textwidth}
    \includegraphics[width=\textwidth]{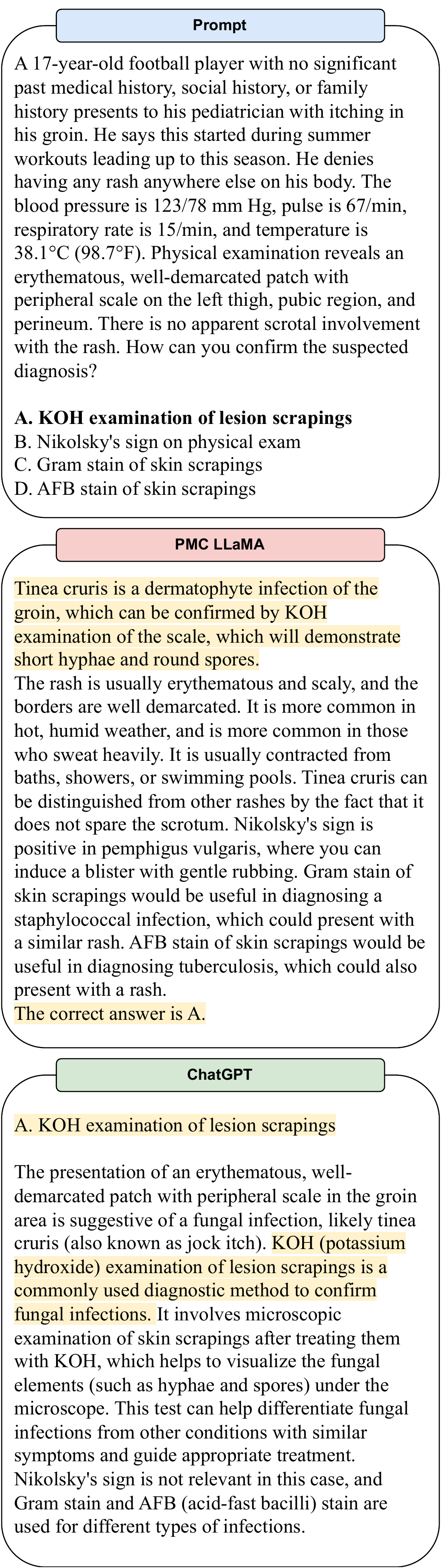}
    \caption{Dermatology}
    \label{fig:case_rationale_2}
\end{subfigure}
\caption{Examples of zero-shot QA rationales covering more medical branches.
The correct options are marked \textbf{bold}.
}
\label{fig:sup_rationale}
\end{figure*}

\vspace{-5pt}
\noindent\textbf{QA Rationales}~
As shown in Fig.~\ref{fig:sup_rationale}, PMC-LLaMA and ChatGPT demonstrate commensurate abilities through diverse medical branches, which is within expectation.
Due to our tons of efforts in knowledge injection, PMC-LLaMA possesses profound knowledge and tends to give thorough explanation of each option in question. This could be a favorable feature to those who trying to have a comprehensive knowledge about patients' symptoms.\\

\vspace{-5pt}
\noindent\textbf{Conversations}~
Though PMC-LLaMA shows high competence on knowledge intensive cases, it's not as good in free-form conversations, especially when the topic branches off from medical domain.\\

\begin{figure*}[htb]
\centering
\begin{subfigure}[][][t]{0.33\textwidth}
    \includegraphics[width=\textwidth]{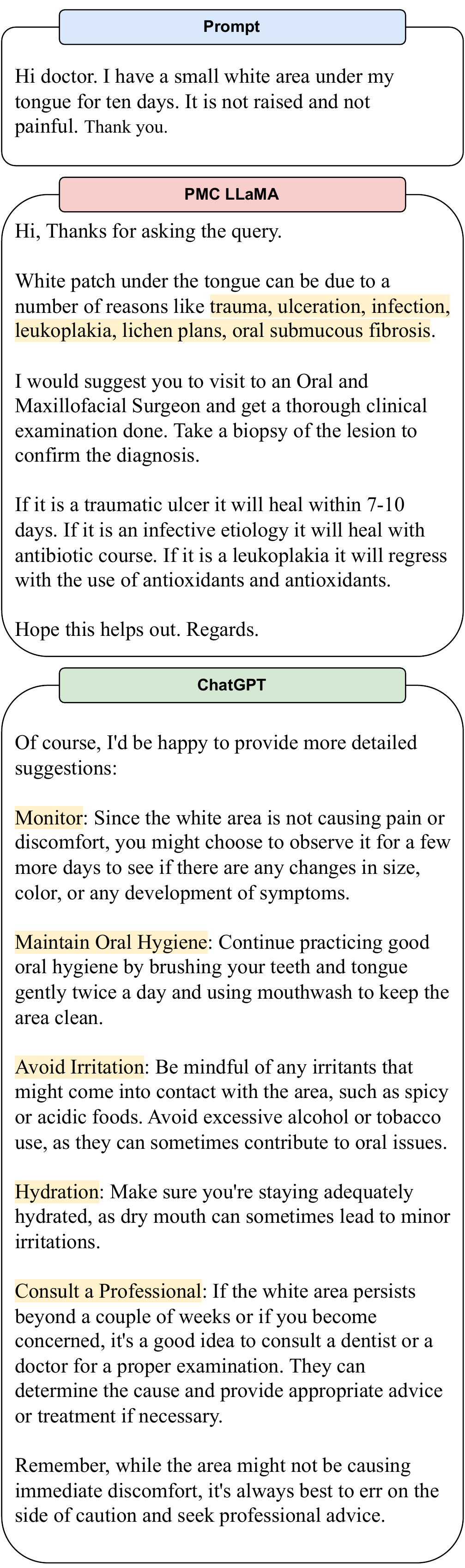}
    \caption{General Health Advice}
    \label{fig:case_chat}
\end{subfigure}
\hfill
\begin{subfigure}[][][t]{0.33\textwidth}
    \includegraphics[width=\textwidth]{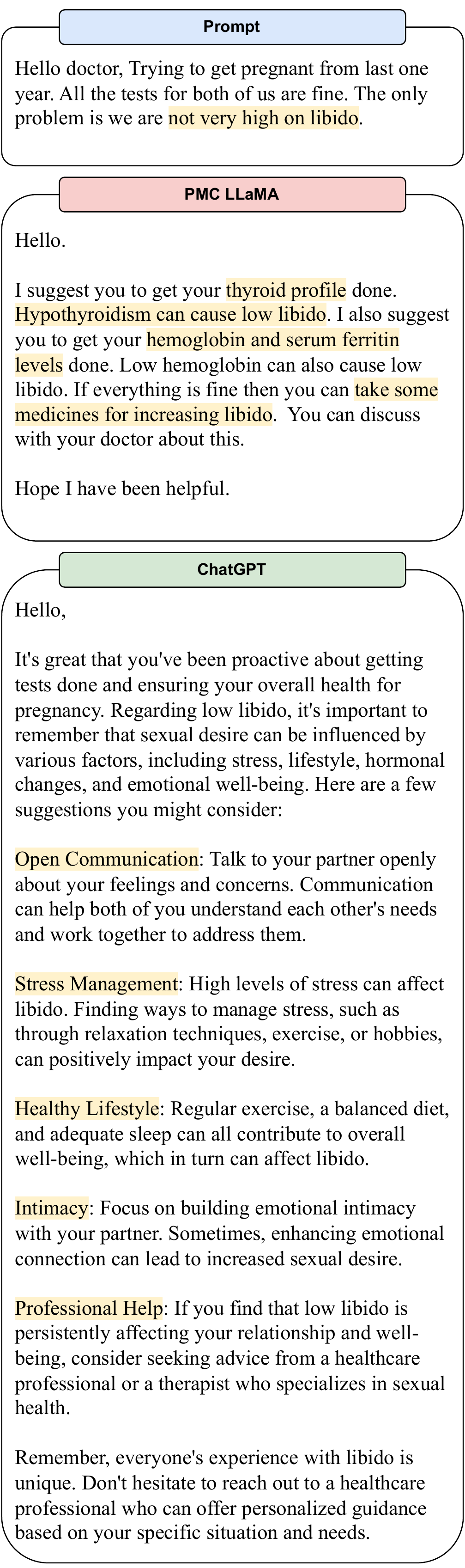}
    \caption{Insufficient Symptomatic Description}
    \label{fig:case_rationale_1}
\end{subfigure}
\hfill
\begin{subfigure}[][][t]{0.33\textwidth}
    \includegraphics[width=\textwidth]{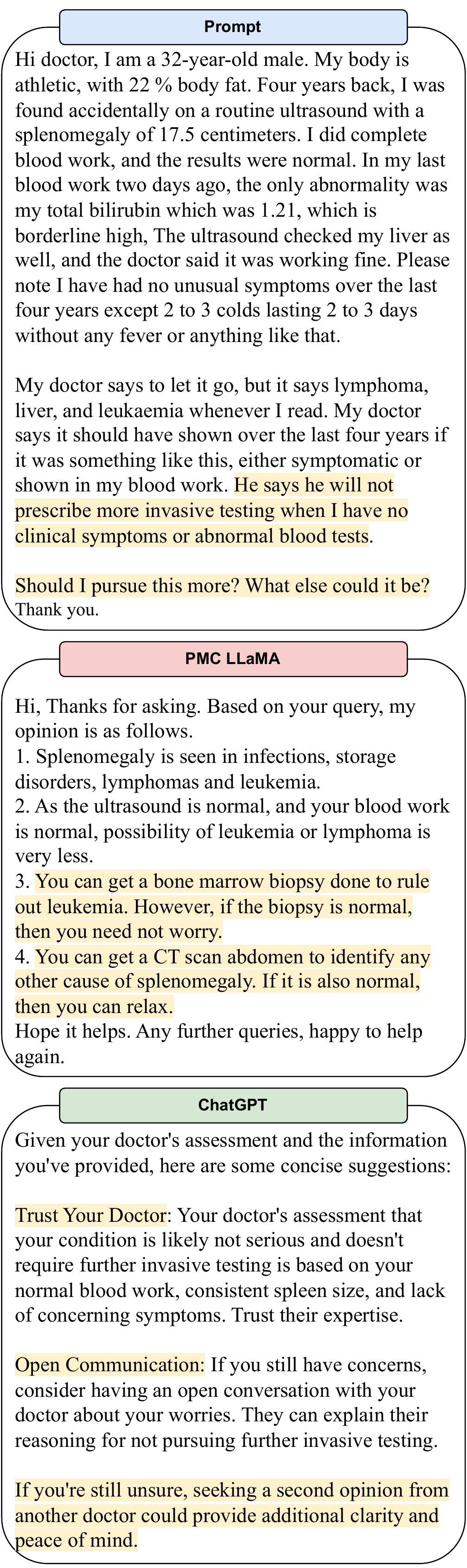}
    \caption{Doctor-Patient Relationship}
    \label{fig:case_rationale_2}
\end{subfigure}
\caption{Examples of zero-shot conversation samples in more complex scenarios. Patient's query could be incomplete or unrelated to disease diagnosis.
}
\label{fig:sup_chat}
\end{figure*}

\vspace{-5pt}
\noindent
In Fig\ref{fig:sup_chat} (a), the patient describes a minor problem under his tongue. ChatGPT suggests the patient to monitor the symptom development, and to consult a professional if necessary. While PMC-LLaMA suggests the cause as rather severe disease, which could be an overestimation.
Also, Fig\ref{fig:sup_chat} (b) shows the user is trying to get pregnant who lacks libido. In this case, the proper advice should start from daily practice instead of prescribed medicine given by PMC-LLaMA.
Fig~\ref{fig:sup_chat} (c) presents a subtle case where the patient has low confidence about his physician's diagnosis and ask for second opinion.
Any negative implication might give misguidance to the patient. Herein PMC-LLaMA gives straight suggestion to help the patient validate his health condition. But ChatGPT's response is wiser. It first attempts to persuade the patient to trust his own physician, and recommends further communication with the doctor.\\

\noindent
Given the broader array of zero-shot samples, we can safely conclude that PMC-LLaMA have powerful abilities on knowledge intensive questions, and further exploration is needed on conversation ability.

\end{document}